\documentclass[10pt,twocolumn,letterpaper]{article}

\usepackage{cvpr}
\usepackage{times}
\usepackage{epsfig}
\usepackage{graphicx}
\usepackage{amsmath}
\usepackage{amssymb}

\usepackage{algorithm}
\usepackage{algorithmic}
\usepackage{multirow}
\usepackage{subfigure}
\usepackage{makecell}
\usepackage{color}
\usepackage{booktabs}

\newcommand{\imp}[2]{$#1_{\uparrow #2}$}
\newcommand{\dec}[2]{$#1_{\downarrow #2}$}

\newcommand{\bimp}[2]{$\mathbf{#1}_{\uparrow \mathbf{#2}}$}

\usepackage[pagebackref=true,breaklinks=true,letterpaper=true,colorlinks,bookmarks=false]{hyperref}

\cvprfinalcopy 

\def\cvprPaperID{1435} 

\ifcvprfinal\fi
\begin{document}

\title{Spectral Feature Transformation for Person Re-identification}

\author{
	Chuanchen Luo$^{1,3}$ \quad 
	Yuntao Chen$^{1,3}$\quad 
	Naiyan Wang$^{2}$ \quad 
	Zhaoxiang Zhang$^{1,3,4}$ \\
	$^{1}$ University of Chinese Academy of Sciences \qquad
	$^{2}$ TuSimple\\
	$^{3}$ Center for Research on Intelligent Perception and Computing, CASIA\\
	$^{4}$Center for Excellence in Brain Science and Intelligence Technology, CAS\\
	$^{1}${\tt\small \{luochuanchen2017, chenyuntao2016, zhaoxiang.zhang\}@ia.ac.cn }\\
	$^{2}${\tt\small winsty@gmail.com} \\
}

\maketitle

\begin{abstract}
	
With the surge of deep learning techniques, the field of person re-identification has witnessed rapid progress in recent years.
Deep learning based methods focus on learning a feature space where samples are clustered compactly according to their corresponding identities.
Most existing methods rely on powerful CNNs to transform the samples individually.
In contrast, we propose to consider the sample relations in the transformation. 
To achieve this goal, we incorporate spectral clustering technique into CNN.
We derive a novel module named Spectral Feature Transformation and seamlessly integrate it into existing CNN pipeline with negligible cost, which makes our method enjoy the best of two worlds.
Empirical studies show that the proposed approach outperforms previous state-of-the-art methods on four public benchmarks by a considerable margin without bells and whistles. 

\end{abstract}

\section{Introduction}
\label{sec:introduction}

Person re-identification (ReID) is an indispensable component in surveillance video analysis. 
Given the probe image, person ReID aims at retrieving images belonging to the same identity across multiple non-overlapping camera views.
Thanks to the emergence of deep learning techniques and large scale datasets~\cite{zheng2015scalable,zheng2017unlabeled,li2014deepreid,wei2018person}, the field of person ReID evolves rapidly. 
Though having achieved much progress, it remains challenging due to drastic pose variation, occlusion and background cluttering.

The success of CNN mainly attributes to its strong power to learn discriminative features.
The goal of representation learning is to pull the samples of the same identity compactly, and push those of different identities far away from each other in the embedding space. 
This already resembles a clustering process. 
On the other hand, the well-known spectral clustering algorithm~\cite{Donath1973Lower} just partitions the data into groups such that data from different groups have very low similarities and data within a group have high similarities.
The goal of these two techniques are identical. 
However, the former one mainly utilizes powerful transformation such as CNN on individual samples, the later one utilizes the relations between samples to transform group of samples. 
These two techniques complement naturally.
Given the close connections between these two techniques, it is natural to explicitly integrate  clustering process into current ReID pipeline to take the relations between samples into account, favorably in an end-to-end manner. Recently, Shen \etal proposed GSRW~\cite{shen2018deep} and SGGNN~\cite{shen2018person} which are closely related to our work. 
The key difference between their works and our work lies in that our work focuses on feature transformation, while theirs' focus on similarity transformation.
As shown in later sections, feature transformation offers not only higher efficiency and simpler implementation, but also better performance. 


In the context of supervised person ReID, the partitions are known since the labels are provided. 
Nevertheless, it is not trivial because the typical spectral clustering method involves eigen-decomposition which is computationally expensive. 
Moreover, the gradient w.r.t. eigen-decomposition is also hard to compute.
To enable efficient computation, we equivalently optimize the transition probability from one subgraph to another on the similarity graph.
Consequently, the module is fully differentiable, and only brings marginal computational cost.
In addition, we can also adopt this method in testing, which can further improve the performance. 
Despite its simplicity, the proposed method improve the performance significantly over strong baselines.

In summary, the contributions of this paper are three folds:
\begin{itemize}
	\item To efficiently optimize group-wise similarities among different identities, we incorporate spectral clustering as a component into deep representation learning. 
	\item We devise a novel Spectral Feature Transformation (SPT) module to implement the spectral clustering in an end-to-end manner. It offers significant performance improvement with negligible overhead.
	\item Extensive experiments on four public benchmarks validate the effectiveness of the proposed method. Our approach outperforms other state-of-the-art methods dramatically without any bells and whistles.
\end{itemize}

\section{Related Works}
\label{sec:related}

\noindent
{\bf Person re-identification.} 
Facilitated by deep learning techniques, the field of person ReID has witnessed great progress in the last few years.
Recent efforts on deep learning based person ReID can be roughly categorized into two directions.
One is to improve the network architecture for person ReID.
Besides common techniques in CNN such as multi-scale feature aggregation~\cite{qian2017multi} or attention modules~\cite{li2018harmonious,Wang_2018_ECCV}, tailor-made architectures~\cite{sun2017beyond,suh2018part,sarfraz2017pose,wang2018person} for person ReID are also devised. 
Sun \etal~\cite{sun2017beyond} splited the feature map into several horizontal parts and imposed supervision on them directly.
Suh \etal~\cite{suh2018part} employed a sub-network to learn body part feature and fused it with appearance feature via a bilinear-pooling layer.
Sarfraz \etal~\cite{sarfraz2017pose} exploited joint keypoints as additional input and used multiple branches to capture orientation-specific feature.
These methods explicitly consider the structure of human body to alleviate the impact of occlusion or inaccurate detections, thus improve the performance.

The other direction concentrates on developing discriminative loss functions.
There are two dominant streams in this direction.
One is to introduce the classical metric learning into deep learning, such as contrastive loss~\cite{hadsell2006dimensionality} and triplet loss~\cite{schroff2015facenet}.
The convergence and performance of deep metric learning are highly dependent on the sample organization of the mini-batch in training.
Several works improve on it by selecting the most informative pairs~\cite{sohn2016improved,oh2016deep,hermans2017defense}.
Another stream focuses on reducing the intra-class variance and increasing the inter-class margin on classification loss functions.
For example, center loss~\cite{wen2016discriminative} regularizes the distance between each data sample and its corresponding class center; 
large-margin softmax~\cite{liu2016large} and its variants~\cite{liu2017sphereface,wang2018cosface,wang2018additive} enforce various types of margin on the classical softmax loss function.
They all have demonstrated effectiveness in face recognition and person ReID tasks.

In addition to the aforementioned directions, some works~\cite{zheng2017unlabeled,ma2017pose,ma2018disentangled} emerge recently which leverage powerful Generative Adversarial Networks~\cite{goodfellow2014generative} to generate person images for training.  


\noindent
{\bf Re-ranking.}
Re-ranking is a post-processing technique to refine the ranking of retrieval results. 
In essence, re-ranking methods aim at enhancing the original similarity metric by the information of local neighbors. 
Early works~\cite{jegou2007contextual,qin2011hello} tried to explore k-reciprocal nearest neighbors for general image retrieval. 
Recently, Zhong \etal~\cite{zhong2017re} introduced re-ranking technique into ReID task. 
They combined the Jaccard distance of $k$-reciprocal encodings and the Euclidean distance of original features in post-processing.
Along this line, Sarfraz \etal~\cite{sarfraz2017pose} aggregated distances between expanded neighbors of image pairs to reinforce the original pairwise distance.
Moreover, to take advantage of the diversity within a single feature, Yu \etal~\cite{Yu2017Divide} further fused distances between different sub-features.

\noindent
{\bf Spectral clustering.} 
Spectral clustering is a conventional algorithm for data clustering. 
It was pioneered by Donath \etal~\cite{Donath1973Lower} and became popular in the pattern recognition community since some landmark works~\cite{Shi2000Normalized,ng2002spectral,meila2001random,von2007tutorial}. 
It is based on the spectral graph theory and converts the data clustering problem into the graph partition problem.
In contrast to K-Means, spectral clustering makes no assumption on the structure of cluster.
So it can generalize to more complex scenarios like intertwined spirals. 
Some recent works~\cite{hershey2016deep,shaham2018spectralnet,tang2018normalized,wu2018improving} tried to incorporate spectral clustering with deep learning.
Tang \etal~\cite{tang2018normalized} proposed normalized cut loss for weakly supervised segmentation to regularize the connections between pixels.
Wu \etal~\cite{wu2018improving} optimized Neighborhood Component Analysis (NCA) criterion to preserve the neighborhood structure in the semantic space.
Though spectral clustering has been applied extensively, combining it with CNN in person re-identification are still under investigation. 


\noindent
{\bf Graph convolutional networks.} 
GCN breaks the assumption of convolution that computation can only take place within local regions. 
Due to the complementarity, it is an effective way to aggregate global information in current CNN framework.
It was first proposed by Kipf~ \etal~\cite{kipf2016semi} for semi-supervised classification.
Currently, GCN is a rising research direction in computer vision. For example, 
Yan \etal~\cite{yan2018spatial} modeled dynamics of human body skeletons via graph convolutional networks.
Wang \etal~\cite{wang2018non,wang2018videos} exploited GCN and a equivalent view non-local feature aggregation to capture the spatial-temporal relations between convolutional features and object proposals in the video, respectively. 
As mentioned before, two closest works to ours are~\cite{shen2018deep,shen2018person}.
They both applied similarity transformation on the graph to achieve better results. 
More detailed comparisons and discussions of these two methods are presented in Sec.~\ref{sec:discussion}.

\section{Method}
\label{sec:method}


\begin{figure*}[t]
	\centering
	\includegraphics[scale=0.8]{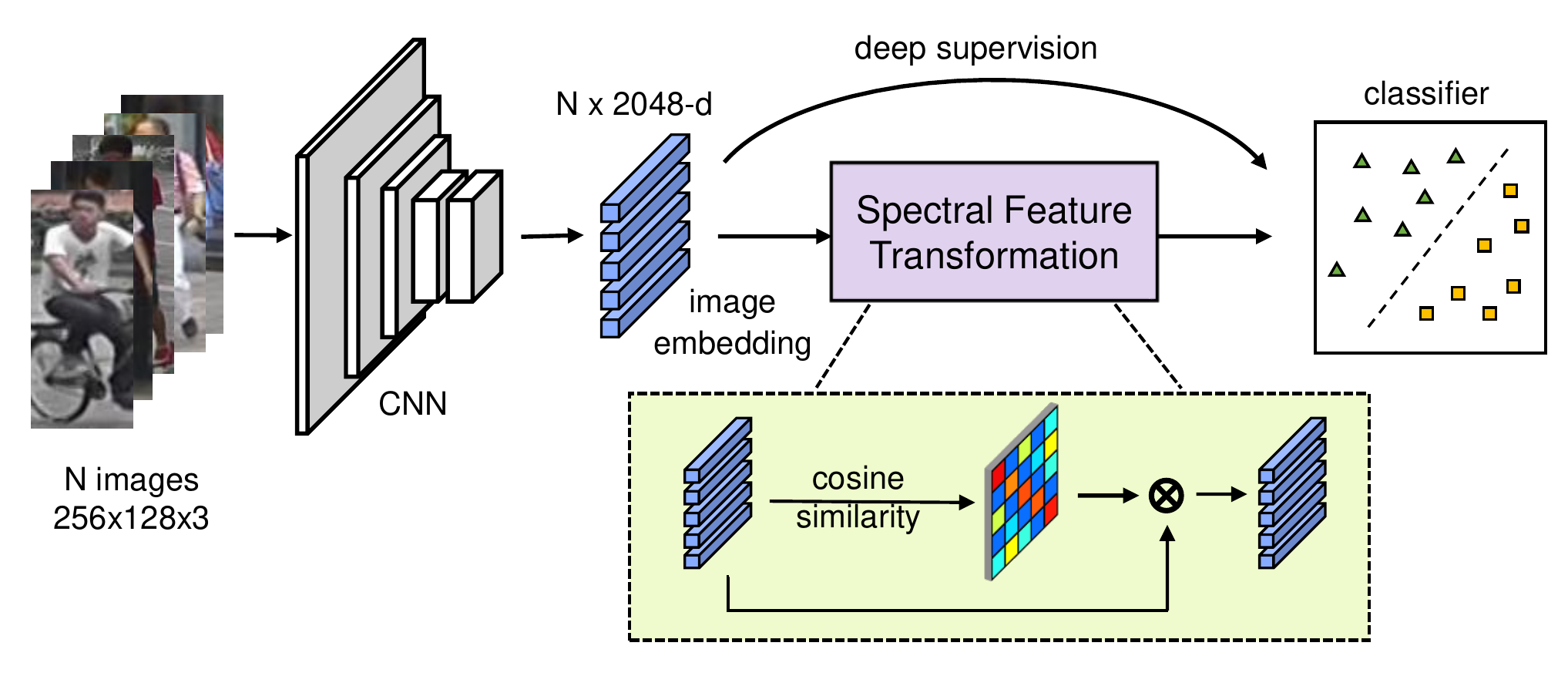}
	\caption{The overall architecture of the proposed model. We adopt the output of global average pooling layer of ResNet as image embedding. The embeddings of training batch then undergo spectral feature transformation. A shared classifier is imposed on the feature before and after the transformation to provide the supervisory signal.}
	\label{fig:model}
\end{figure*}

Given the close connection between clustering and deep learning based person ReID, it is natural to bring clustering techniques into deep person ReID.
Among dozens of candidates, spectral clustering shows superiority in many aspects.
In contrast to k-means which clusters data points by Euclidean distance, spectral clustering focuses on the set of similarities between and within groups which is more flexible. It makes no prior assumption on data topology, thus makes it applicable to more complex scenarios. 



We first give a brief introduction of spectral clustering algorithm in Section~\ref{sec:spectral}.
We then elaborate the proposed Spectral Feature Transformation (SFT) in Section~\ref{sec:sft}.
In Section~\ref{sec:training} and ~\ref{sec:offline}, we explain the training strategies in details and then extend the proposed SFT module to the post-processing stage.

\subsection{Graph Cut and Spectral Clustering}
\label{sec:spectral}
We first review the classical spectral clustering and its closely related concept graph cut.
From the viewpoint of graph, data $X=\{x_i\}_{i=1,...,n}$ can be represented as a undirected graph. 
Wherein, each data point in $X$ is a vertex of the graph and the edge is weighted by the similarity of corresponding data points $w_{ij}=\mathrm{sim}\left(x_i,x_j\right)$.
For brevity, we take the 2-cluster problem as an example in the following formulation, and readers can refer to \cite{stella2003multiclass} for multi-cluster extension.

To obtain the optimal clustering result on a graph, an intuitive way is to solve a minimum cut problem.
For two disjoint subsets $A,B\subset X$, the cut between them is defined as 
\begin{equation}
\mathrm{cut}(A,B)=\sum_{i \in A,j\in B} w_{ij} .
\end{equation}

However, minimizing vanilla cuts often leads to a trivial solution where a single vertex is separated from the rest of the graph. 
To circumvent the issue, Shi \etal~\cite{Shi2000Normalized} proposed to normalize each subgraph by its volume:
\begin{equation}
\mathrm{Ncut}(A,B)=\frac{\mathrm{cut}(A,B)}{\mathrm{vol}(A)}+\frac{\mathrm{cut}(A, B)}{\mathrm{vol}(B)},
\end{equation}
where $\mathrm{vol}(A)=\sum_{i\in A, j\in X} w_{ij}$ is the total connection from nodes in $A$ to all nodes in the graph.


\subsection{Spectral Feature Transformation}
\label{sec:sft}


%

Suppose $X\in \mathbb{R}^{n \times d}$ is the final embedding of a training batch.
Wherein, $n$ and $d$ denote the number of data points and the dimension of the embedding vector, respectively.  
We adopt cosine similarity with exponential transformation to measure the affinities between samples. 
Formally, each element of affinity matrix $W$ is defined as  
\begin{equation}
w_{ij} = \mathrm{exp}\left( \frac{x_i^Tx_j}{\sigma\cdot\Vert x_i\Vert_2\Vert x_j\Vert_2}\right) ,
\label{eq:affinity}
\end{equation}
where $\sigma$ is the temperature parameter. Now, we can define a graph on the samples in a mini-batch as $G = (X, W)$.
If the affinity is properly normalized, it actually defines a random walk process on the graph.
By normalizing the rows of $W$ to 1, we can derive the stochastic matrix $T$, which reflects the transition probability from nodes to nodes: 
\begin{equation}
\label{eqn:transition}
T=D^{-1}W,
\end{equation}
where $D$ is a diagonal matrix whose elements are defined as $d_i =\sum_{j=1}^n w_{ij}$.
In practice, the computation of $T$ can be implemented by applying softmax function with temperature $\sigma$ on affinity matrix $W$ .

The most intriguing property of $T$ is the escaping probability $P(A\to\bar{A})$ which measures the transition probability from a subgraph $A \subset X$ to another subgraph $\bar{A} = X - A$. For ReID task, a subgraph $A$ denotes the set of samples belonging to the same identity, and the escaping probability is essentially the chance of an identity getting misclassified. We will elaborate the formal definition of $P(A\to\bar{A})$ below.

The stationary distribution of the underlying random walk process is simply given as,
\begin{equation}
\label{eqn:stationary}
\pi_i = \dfrac{d_i}{\mathrm{vol}(X)},
\end{equation}
$\pi_i$ represents the normalized connection strength of one sample to the rest of the graph. A sample with more similar samples within graph tends to have larger connection strength. 

And finally, combining Eqn.~\ref{eqn:transition} and ~\ref{eqn:stationary}, the escaping probability is defined as,
\begin{equation}
P(A\to \bar{A}) = \dfrac{\sum_{i\in A, j\in\bar{A}}\pi_i T_{ij}} {\sum_{i\in A}\pi_i}.
\end{equation}
It is straightforward that a small $P(A\to \bar{A})$ requires strong intra-cluster connections and weak inter-cluster connections, which is the desired property for spectral clustering. In fact, as shown in \cite{meila2001random}, 
\begin{equation}
\mathrm{Ncut}(A,\bar{A})=P(A\to\bar{A})+P(\bar{A}\to A).
\end{equation}

In the fully supervised setting, the partition $A$, $\bar{A}$ is given, so we mainly focus on minimizing the transition probability $P(A\to \bar{A})$ w.r.t. the stochastic matrix $T$. By minimizing the transition probability, we essentially minimize the probability of misclassifying a data sample in group $A$ into group $\bar{A}$. Note that this supervision is applied in group-wise not image-wise, which is fundamentally different from previous works. We can directly utilize the cross entropy loss w.r.t. one-hot ground truth label to optimize our clustering objective function. 

This can be simply implemented by multiplying $T$ with original feature $X$, and optimizing the spectral feature transformation module with standard SGD in an end-to-end manner. 
The overall architecture of the proposed neural networks is displayed in Figure~\ref{fig:model}.



\subsection{Training Strategy}
\label{sec:training}
In the early stage of training, the features extracted by neural networks are too ambiguous to describe corresponding images accurately.
Transition probabilities derived from these features are thus not reliable.
This drifts feature transformation and makes training process unstable.
Though warming up strategy~\cite{goyal2017accurate} mitigates the issue to some extent, the performance is still not satisfactory on some datasets.  
To eliminate the problem, we impose supervision directly on original features to prevent them from drifting. 
We draw this inspiration from deeply-supervised nets~\cite{lee2015deeply}.
It is noteworthy that the original feature and the transformed feature share the same classifier during training.
Only by then can we guarantee that the modes of features are aligned before and after spectral feature transformation.

To fully liberate the power of spectral clustering, it is necessary to satisfy the assumption that the input data obey the underlying cluster structure.
In other words, there must be sufficient images for each identity in a training batch. 
Thus, we adopt the sampling strategy proposed by Hermans \etal~\cite{hermans2017defense} which is ubiquitous in deep metric learning.
Specifically, a mini-batch in training contains $P$ identities and each identity has $K$ images.

\subsection{Post-processing}
\label{sec:offline}
Inspired by works~\cite{shen2018deep,shen2018person}, we further extend the proposed spectral feature transformation to the post-processing stage.
The extension is based on the assumption that there is an underlying cluster structure in the neighborhood of the probe image in the gallery.
As the evaluation protocol implies, top ranking list has a larger impact on the final performance.
So we only refine the top-$n$ ranking list to balance between efficiency and performance.

Given a probe image, images in the gallery are ranked according to their cosine similarities with it.
Then, we collect features of top-$n$ items and perform spectral feature transformation on them.
In the end, top-$n$ rank list is recomputed based on the transformed features. 
Since $n$ is much smaller than the size of gallery and the features are extracted in advance, the refinement process introduces negligible overhead. 


\section{Discussion}
\label{sec:discussion}
In the sequel, we will analyze some appealing properties of our method which contribute to the improvement.
Next, distinctions and connections with other works will also be discussed. 

\subsection{Appealing Properties}
\noindent
{\bf Relax Assumptions and Ease Optimization}
Instead of directly constraining the pairwise similarities, our method relaxes the learning objective to optimize such similarities after the group-wise transformation by SFT. The transformation introduced by SFT moves the features towards the cluster center, thus it has enhanced the discrimination of features.
This makes the optimization and feature learning easier and finally leads to higher performance. \\


\noindent
{\bf Training Diversity}
According to the definition, all samples in the mini-batch participate in the SFT operation.
The transformed feature of one data sample may differ because the composition of the mini-batch changes in each training epoch.
This desired property introduces massive diversity which effectively alleviates the risk of over-fitting.

\subsection{Distinctions with SGGNN}
SGGNN~\cite{shen2018person} addressed the ReID problem from the viewpoint of graph which is similar to us.
However, there are some obvious discrepancies.
In terms of the definition of the graph, for each image in the probe set, they construct one graph with probe-to-gallery similarities as nodes and gallery-to-gallery similarities as edges.
While in our approach, each node directly corresponds to the feature of a sample and each edge is defined as the similarity of its endpoints. 
Consequently, in each mini-batch, they need to construct $P$ graphs with $P\times\left( K-1 \right)$ nodes, while we treat the whole mini-batch as one single graph which is much conceptually  simpler and faster.

\subsection{Distinctions with Ncut Loss}
Both Ncut loss and our method focus on optimizing a group-wise similarities.
However, the detailed implementations are different. 
Ncut loss realizes the goal via directly optimizing Rayleigh quotient.
It is equivalent to minimize the transition probability as previously stated.
It is often used along with a extra cross entropy loss to constrain the learning of the feature.
We will compare the two loss functions in Section.~\ref{sec:ablation}. 

\subsection{Distinctions with Graph Convolution}
Graph convolution is proposed to relax the spatial connectivity assumptions in traditional convolution operator. In particular, instead of computing on the spatially connected locations, graph convolution computes on the whole graph weighted by the connectivity of each node. It generalizes convolution operator on non-Euclidean data.
On the contrary, the proposed SFT module is a non-parametric operation applied on the final feature to enhance the features. 
These two methods differ in their motivations. 

\section{Experiments}
\label{sec:experiment}
In this section, We conduct extensive experiments on four public person re-identification benchmarks, \ie, Market-1501~\cite{zheng2015scalable}, DukeMTMC-ReID~\cite{zheng2017unlabeled,ristani2016MTMC}, CUHK03~\cite{li2014deepreid} and MSMT17~\cite{wei2018person}. 






\subsection{Evaluation Protocols} 
We follow the standard ReID evaluation metric to evaluate our method and compare it with other works.
Given a query image, gallery images are ranked according to their cosine similarity with it. 
Based on the generated ranking list, Cumulated Matching Characteristics (CMC) at rank-1, rank-5 and mean average precision (mAP) are calculated to evaluate the performance of the model. 

Note that on CUHK03 dataset, we use the recently proposed protocol in~\cite{zhong2017re}. 
The new protocol splits the whole dataset into 767 and 700 identities for training and testing, respectively, which is much harder than the original one.
On Market-1501, besides the traditional single query protocol, we also introduce 500k distractors to evaluate the scalability and robustness of the proposed ReID model.

\subsection{Implementation Details}
We adopt ResNet-50~\cite{he2016deep} pre-trained on ImageNet~\cite{deng2009imagenet} as our backbone network. 
We use the output of global average pooling layer of ResNet as the embedding vector.
In order to preserve more fine-grained information, the downsampling of the last stage of ResNet is discarded which leads to a total stride of 16.
The temperature $\sigma$ of SFT layer is set to 0.02 for MSMT17 dataset and 0.1 for the remaining three datasets according to cross validation. 
As for the classifier, we follow a bottleneck design~\cite{sun2017beyond} which has been proven effective by many works.
Specifically, a fully-connected layer is applied to reduce the dimension of the feature from 2048 to 512 which is followed by Batch Normalization~\cite{ioffe2015batch} and PReLU~\cite{He2015Delving}. 
The output is then $l_2$-normalized and fed into the loss function. 
We adopt AM-Softmax~\cite{wang2018additive} loss for the final classification.
In all experiments, the margin and the scaling parameter of AM-Softmax are set to 0.3 and 15, respectively.

In terms of data pre-processing, input images are resized into $256 \times 128$. 
Random horizontal flipping and random erasing~\cite{zhong2017random} are utilized as data augmentation. 
In training, each mini-batch contains 16 persons and each person has 8 images and results in a batch size of 128. 
Stochastic Gradient Descent (SGD) with the momentum of 0.9 is applied for optimization.
We train 140 epochs in total. 
The learning rate warms up from 0.001 to 0.1 linearly in the first 20 epochs. 
It is decayed to 0.01 and 0.001 at 80th and 100th epoch, respectively. 
In the post-processing stage, we refine the top-50 ranking list for each probe image on Market-1501, DukeMTMC-ReID and CUHK03.
While for MSMT17, top-150 ranking list is refined, since it has a much larger gallery than the other datasets.
Our implementation is based on MXNet~\cite{chen2015mxnet} framework. 
The codes will be public available later.

\subsection{Ablation Study}
\label{sec:ablation}

\begin{table*}
	\begin{center}
		\resizebox{\textwidth}{!}{
		\begin{tabular}{ccccc|cc|cc|cc|cc}
			\Xhline{1.1px}
			\multicolumn{5}{c|}{Component} & \multicolumn{2}{c|}{Market-1501} & \multicolumn{2}{c|}{DukeMTMC}  & \multicolumn{2}{c|}{CUHK03} & \multicolumn{2}{c}{MSMT17} \\
			\hline
			sft & ds(u) & ds(s) & post & kr & mAP & Rank-1  & mAP & Rank-1  & mAP & Rank-1 & mAP & Rank-1\\
			\hline
			& & & & & 77.3 & 91.2 & 66.1 & 83.3 & 40.6 & 44.9& 37.3 & 66.7  \\
			\checkmark & & & & & \imp{79.6}{~2.3} & \imp{91.6}{0.4}  & \imp{70.4}{~4.3} & \imp{85.4}{2.1}  & \imp{60.2}{19.6} & \imp{66.3}{21.4}  & \imp{44.7}{~7.4} & \imp{71.9}{~5.2}\\
			\checkmark & \checkmark & & & & \imp{79.0}{~1.7} & \imp{91.8}{0.6}  & \imp{66.9}{~0.8}  & \imp{83.3}{0.0} & \imp{43.1}{~2.5} & \imp{47.1}{~2.2} & \dec{35.3}{~2.0} & \dec{63.7}{~3.0}  \\
			\checkmark & & \checkmark & & & \imp{82.7}{~5.4} & \imp{93.4}{2.2} & \imp{73.2}{~7.1} & \imp{86.9}{3.6} & \imp{62.4}{21.8} & \imp{68.2}{23.3} & \imp{47.6}{10.3} & \imp{73.6}{~6.9}  \\
			\checkmark & & \checkmark & \checkmark & & \imp{87.5}{10.2} & \bimp{94.1}{2.9}  & \imp{79.6}{13.5} & \bimp{90.0}{6.7} & \bimp{71.7}{31.1} & \bimp{74.3}{29.4} & \imp{58.3}{21.0} & \bimp{79.0}{12.3} \\
			\checkmark & & \checkmark &  & \checkmark & \bimp{90.6}{13.3} & \imp{93.5}{2.3} & \bimp{83.3}{17.2} & \imp{88.3}{5.0} & \imp{68.7}{28.1} & \imp{71.7}{26.8} & \bimp{60.8}{23.5}  & \imp{76.1}{10.6}  \\
			\Xhline{1.1px}
		\end{tabular}}
	\end{center}
	\caption{Ablation studies on Market-1501, DukeMTMC-ReID, CUHK03(labeled) and MSMT17 dataset. \textbf{sft} represents the proposed spectral feature transformation. \textbf{ds(s)} and \textbf{ds(u)} denote deep supervision on the original feature with shared and unshared classifier, respectively. \textbf{post} means the proposed post-processing operation. \textbf{kr} denotes k-reciprocal encoding re-ranking method.}
	\label{tab:ablation}
\end{table*}


To investigate the effect of each component we proposed, a series of controlled experiments are conducted on all benchmarks we mentioned above.   

\begin{figure}[h]
	\centering
	\subfigure[Without SFT]{
		\label{fig:heat:a}
		\includegraphics[scale=0.138]{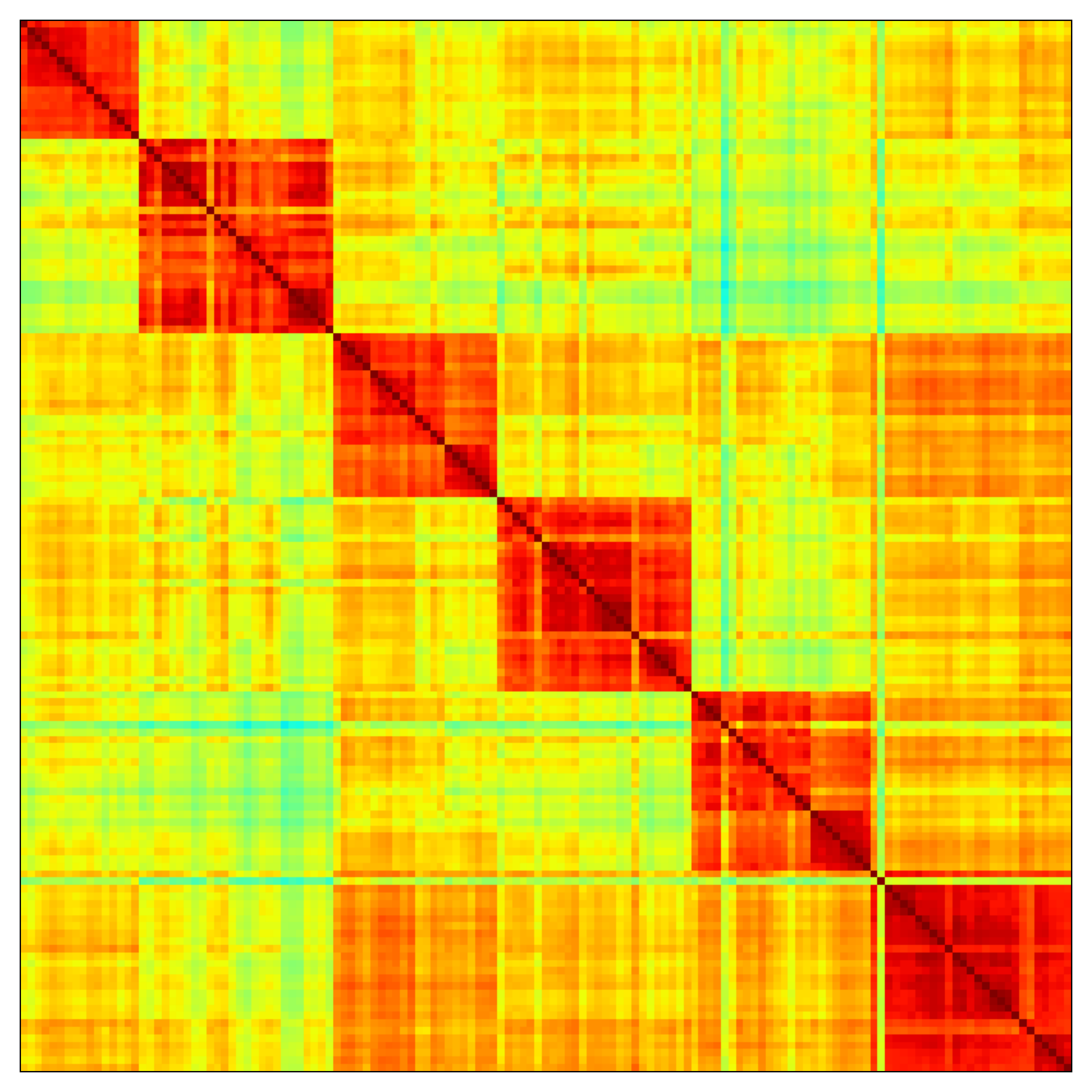}
	}
	\subfigure[With SFT]{
		\label{fig:heat:b}
		\includegraphics[scale=0.138]{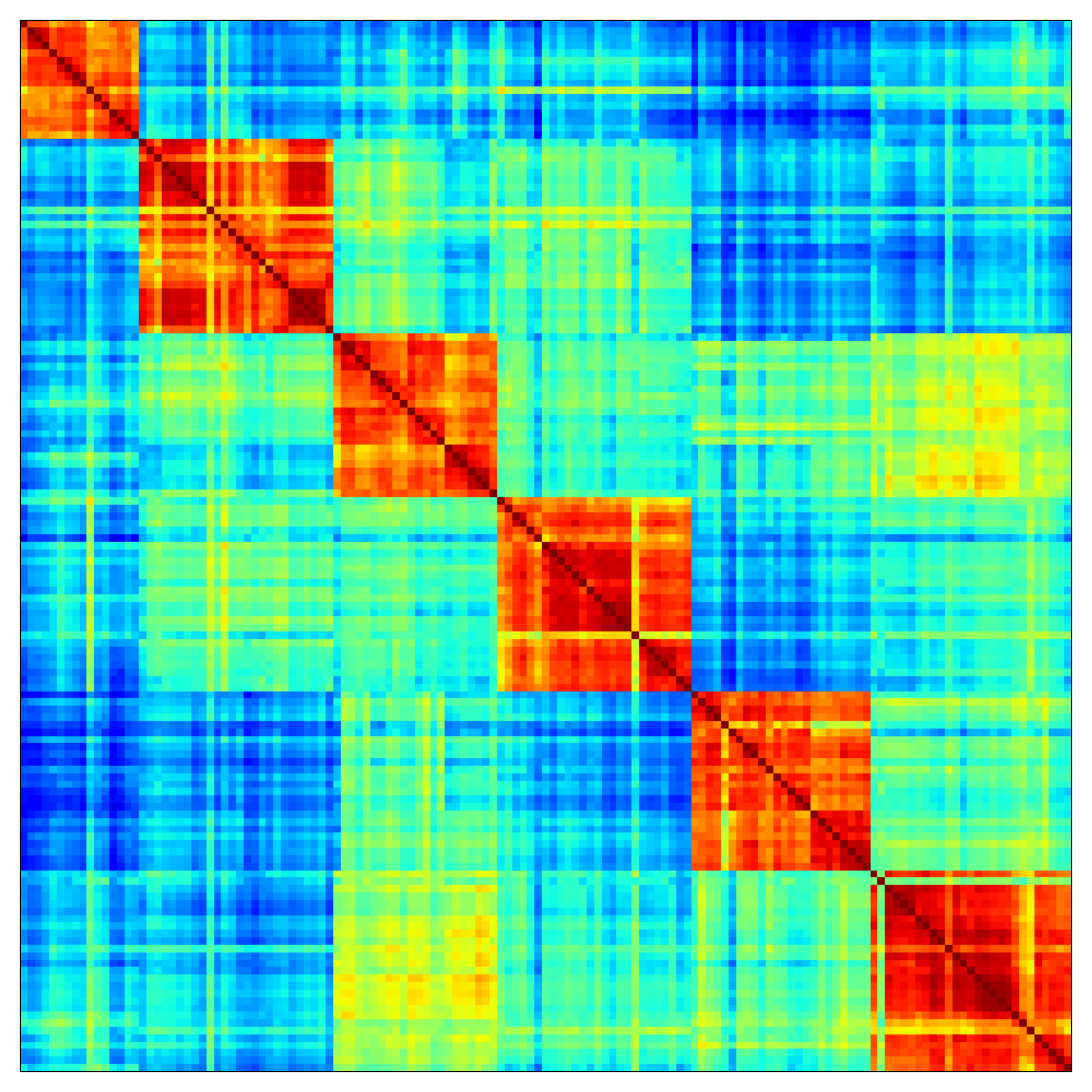}
	}
	\subfigure{
	\includegraphics[height=34.5mm]{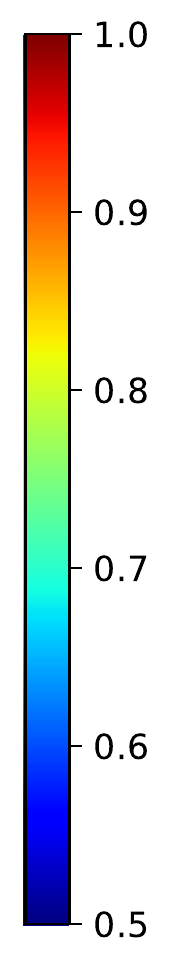}
}
	\caption{Visualization of the affinity matrix. We randomly sample 6 identities from DukeMTMC-ReID and take all images belonging to them for visualization. For clarification, samples are arranged according to their identities. It can be seen that the proposed spectral feature transformation obviously suppresses the similarities among different identities.}
	\label{fig:heat}
\end{figure}

\noindent
{\bf Effectiveness of Spectral Feature Transformation.}
As shown in the first two rows of Table~\ref{tab:ablation}, consistent improvements are achieved on all four benchmarks. 
Rank-1 accuracy/mAP are improved by 0.4\%/2.3\%, 2.1\%/4.3\%, 21.4\%/19.6\%, 5.2\%/7.4\% on Market-1501, DukeMTMC-ReID, CUHK03 and MSMT17, respectively. 
Our method is advantageous in the dataset where each identity is consisted of abundant views as CUHK03, DukeMTMC-ReID and MSMT17. 
In addition, we visualize the affinity matrix of images of 6 different identities extracted with and without SFT module.
It can be observed from Figure~\ref{fig:heat} that the connections between different identities are obviously suppressed.
Thus, the features extracted by our method are more discriminative for person ReID.

\noindent
{\bf Effectiveness of Deep Supervision.} 
To investigate the influence of deep supervision mentioned in Section~\ref{sec:training}, we combine it with our SFT module while training.
It can be seen from row 1 and row 4 in Table~\ref{tab:ablation} that deep supervision contributes to significant improvement. 
Furthermore, we investigate the necessity of sharing classifier for features before and after the module. 
As shown in rows 3-4 of Table~\ref{tab:ablation}, the performance drops back to baseline level when independent classifier is applied for training. 
This indicates the classifier for the original feature could dominate the training process in the unshared setting.

\begin{figure}[h]
	\centering
	\includegraphics[scale=0.45]{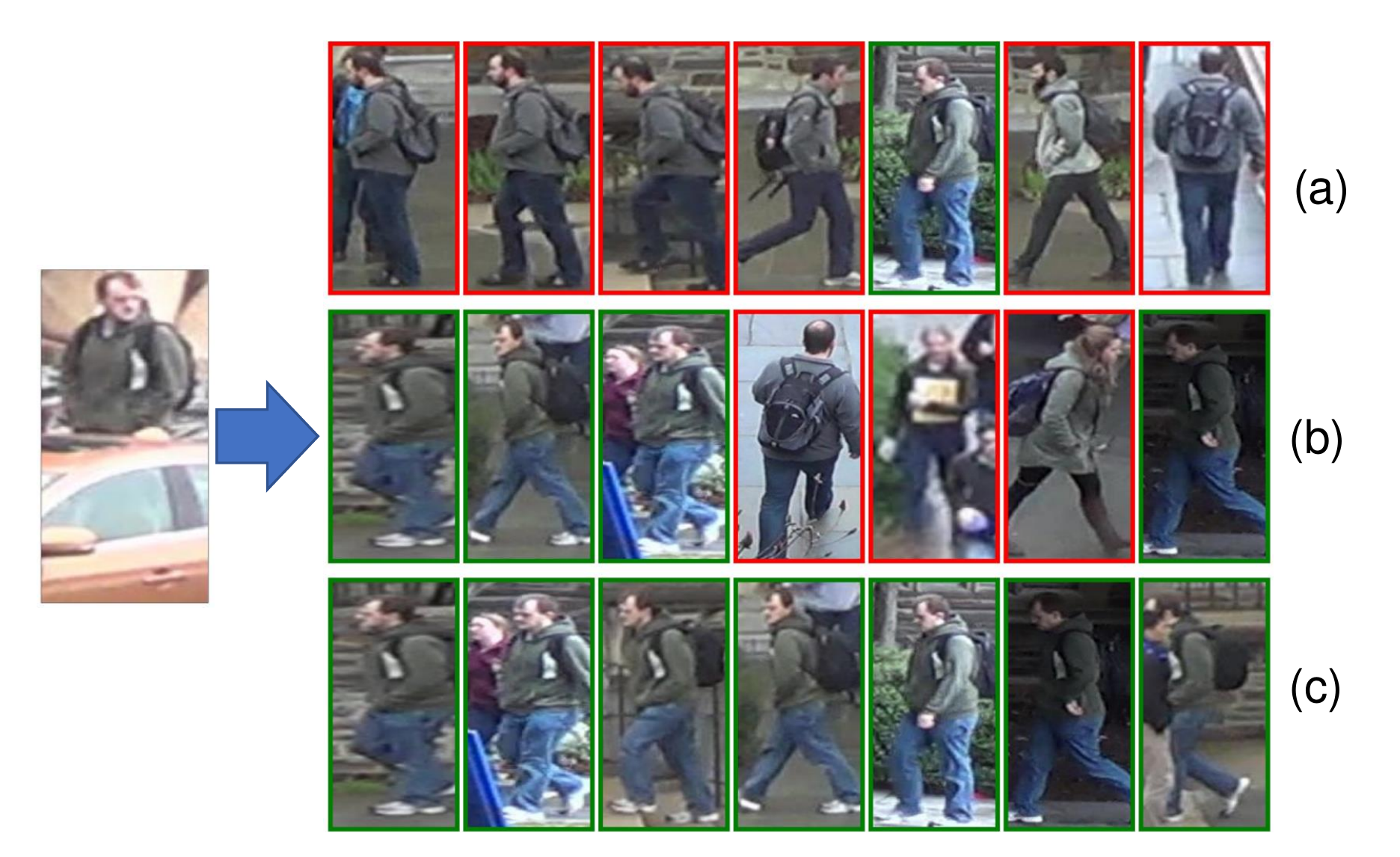}
	\caption{An example of the retrieval. (a) is the result of the baseline. (b) is generated by our method. (c) is the refined result of (b) by the proposed post-processing method.}
	\label{fig:retrival}
\end{figure}

\noindent
{\bf Effectiveness of Post-processing.}
We also evaluate our method with and without the proposed post-processing.  
As reported in rows 4-5 of Table~\ref{tab:ablation}, the proposed post-processing could further improves the performance. 
To further clarify its effectiveness, we make a comparison with the $k$-reciprocal encoding~\cite{zhong2017re} method.
First, it is clearly shown that our method can also be compatible with other post-processing method, $k$-reciprocal encoding could still boost the performance significantly.
Moreover, the proposed post-processing surpasses $k$-reciprocal encoding in terms of Rank-1 accuracy which is the most considered metric on all the four datasets. 
For mAP, our post-processing method demonstrates advantages on the CUHK03 dataset, while $k$-reciprocal encoding is better on the other three. 
Compared with the proposed post-processing, $k$-reciprocal encoding uses more neighborhood information which makes it computationally expensive.
Combining all these components, the performance of our method improves dramatically with negligible overhead and minor modification.
An example of the retrieval is represented in Figure~\ref{fig:retrival}.

\begin{figure}[h]
	\centering
	\subfigure[Market-1501]{
		\label{fig:sigma:a}
		\includegraphics[scale=0.323]{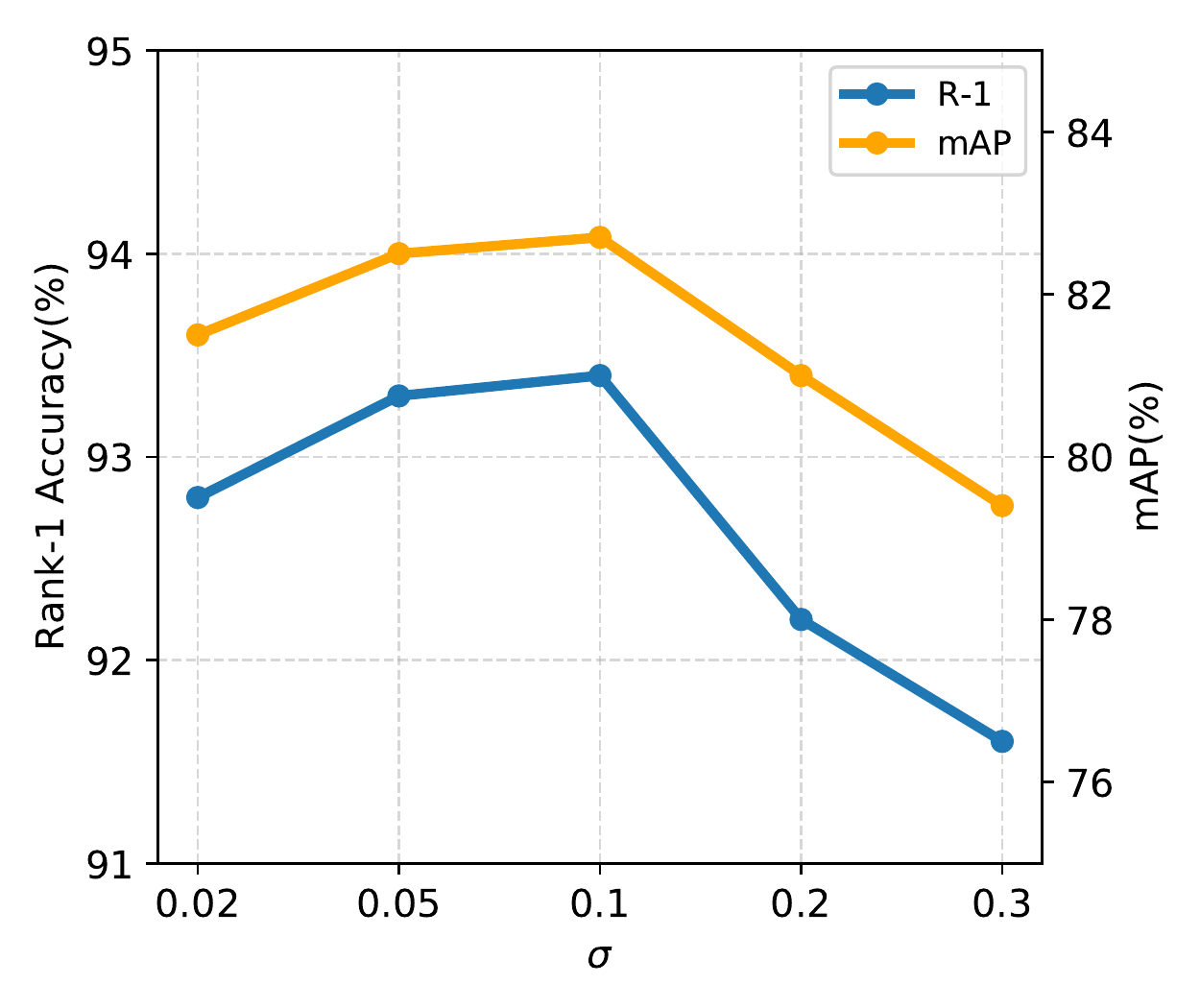}
	}
	\subfigure[DukeMTMC-ReID]{
		\label{fig:sigma:b}
		\includegraphics[scale=0.324]{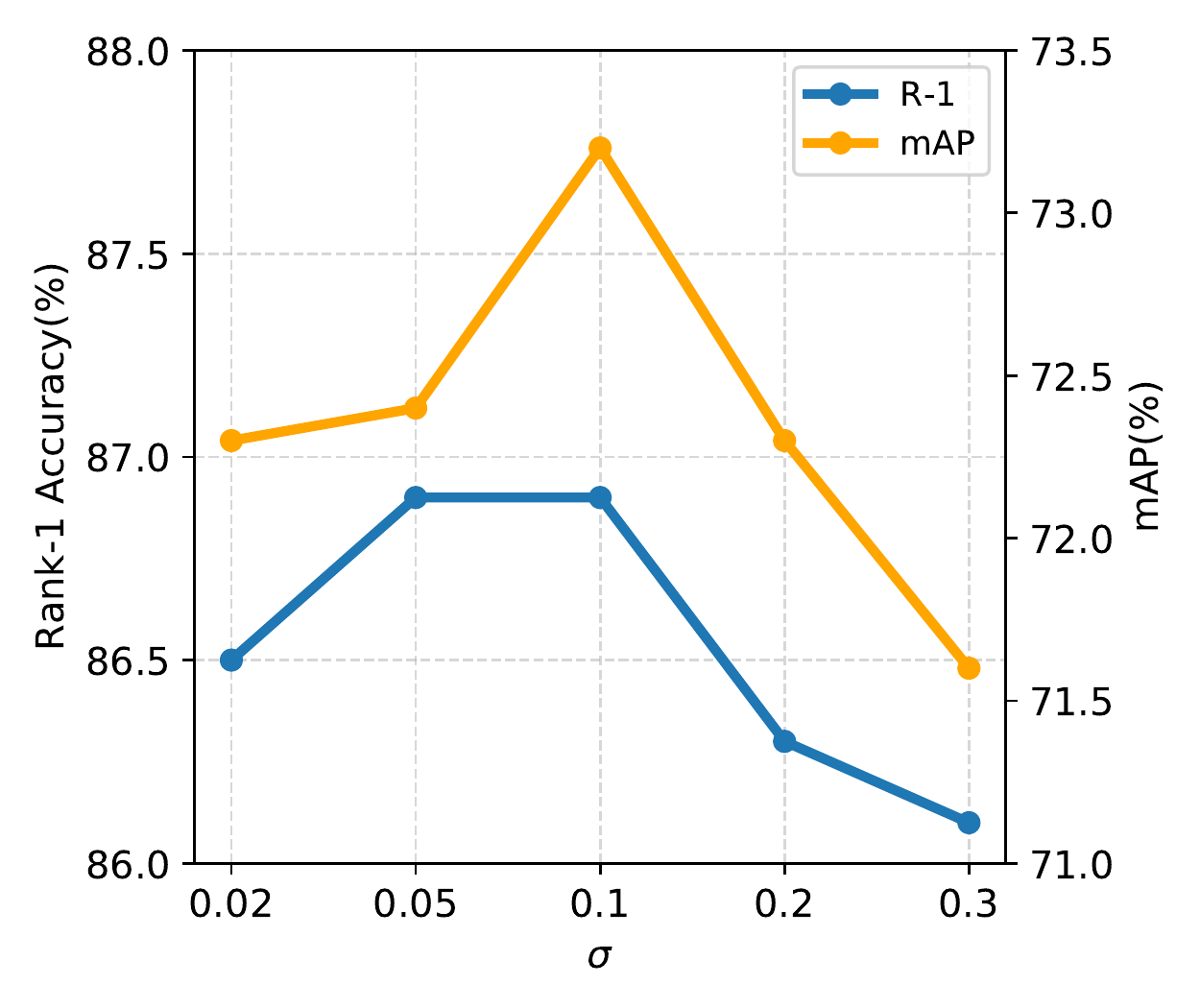}
	}
	\subfigure[CUHK03]{
		\label{fig:sigma:c}
		\includegraphics[scale=0.323]{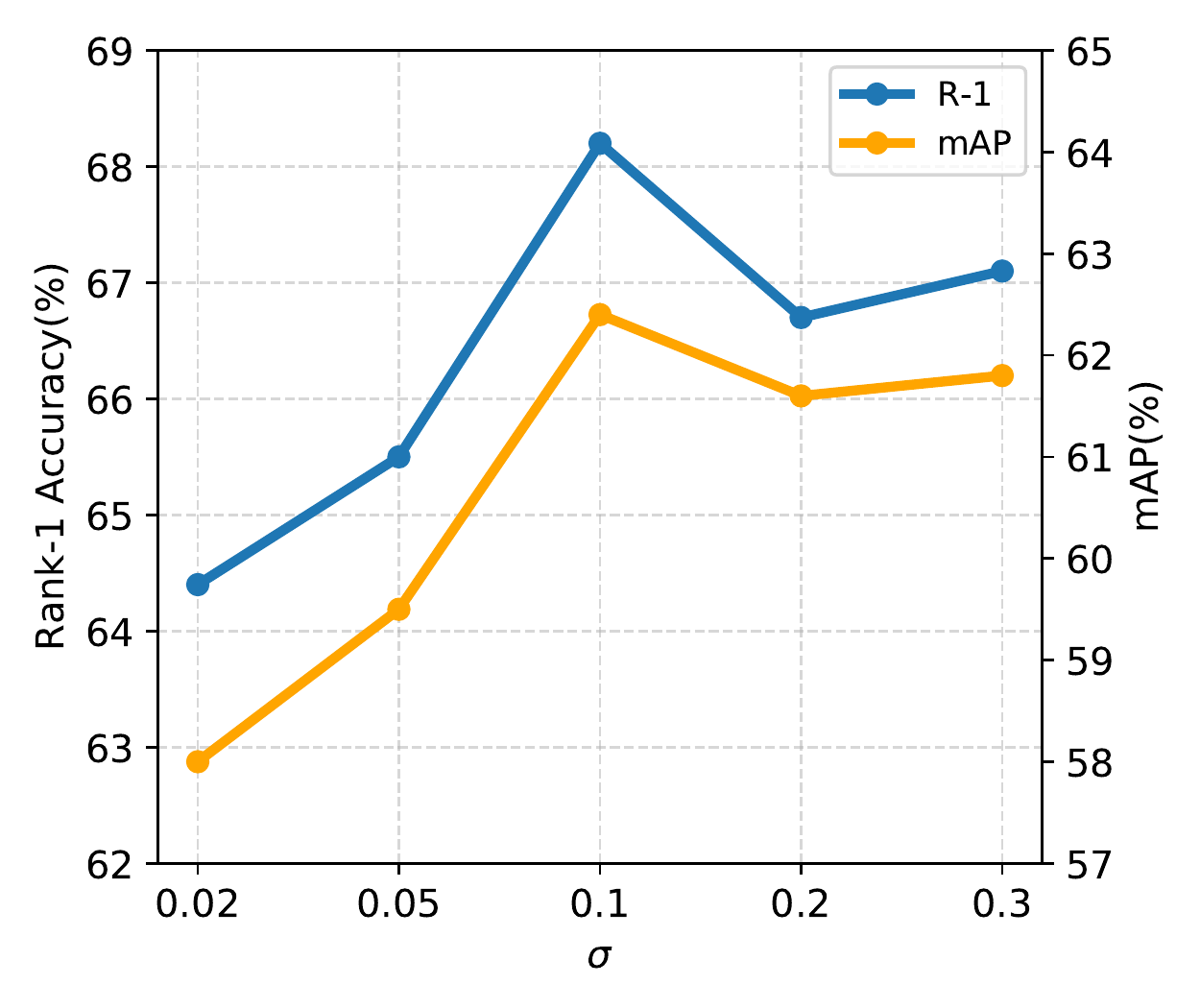}
	}
	\subfigure[MSMT7]{
		\label{fig:sigma:d}
		\includegraphics[scale=0.324]{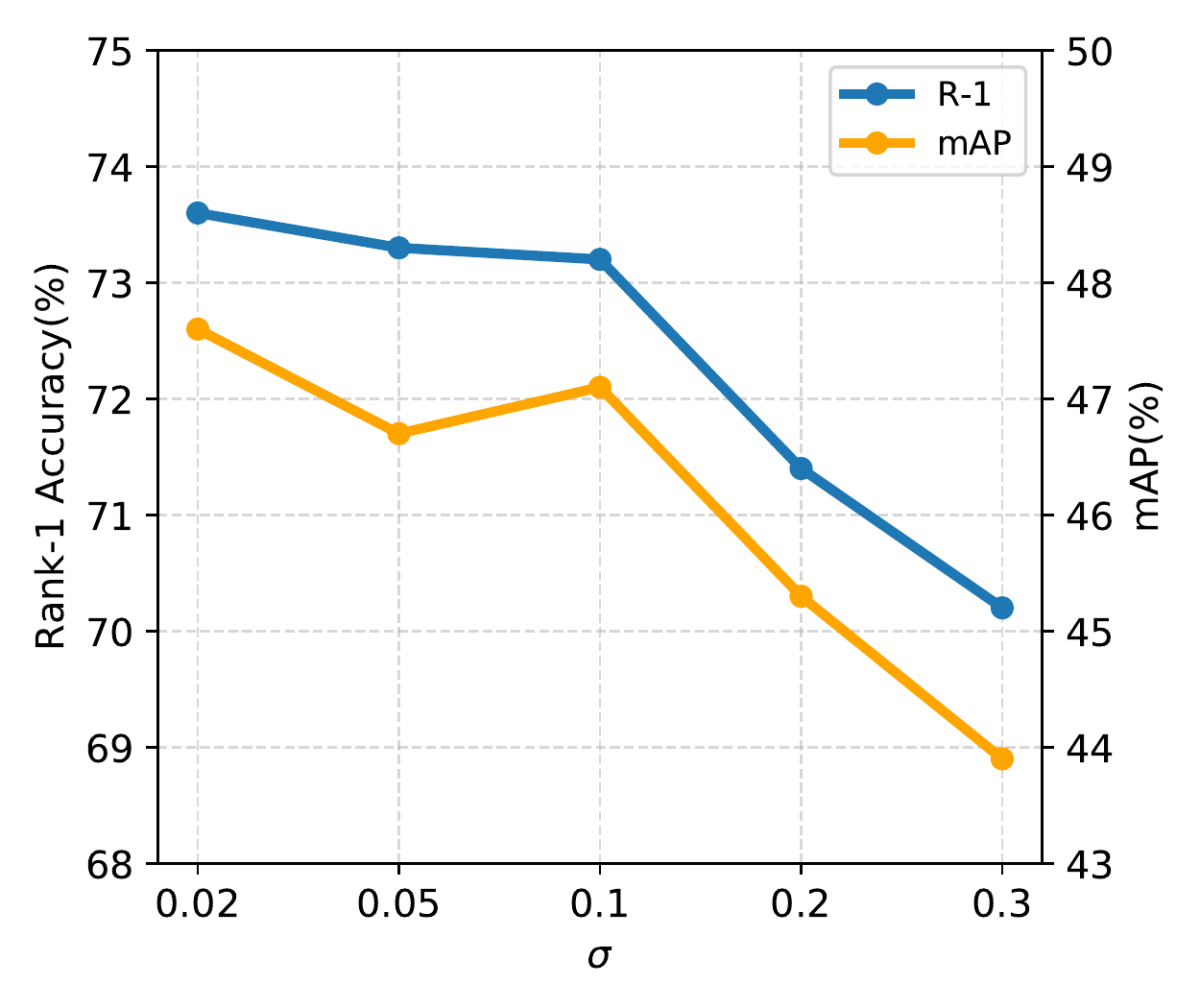}
	}
	\caption{The influence of bandwidth $\sigma$. }
	\label{fig:sigma}
\end{figure}

\noindent
{\bf Influence of Temperature $\sigma$.}
The proper selection of affinity function is crucial for the success of spectral clustering.
So, it is necessary to investigate the impact of $\sigma$ on the learned features. 
To this end, we vary $\sigma$ to five different values and evaluate the performance of the model trained under these settings.
The results are visualized in Figure~\ref{fig:sigma}. 
It shows that our method is relatively robust to the value of $\lambda$. 
%

\begin{figure}[h]
	\centering
	\subfigure[DukeMTMC-ReID]{
		\label{fig:batch:a}
		\includegraphics[scale=0.345]{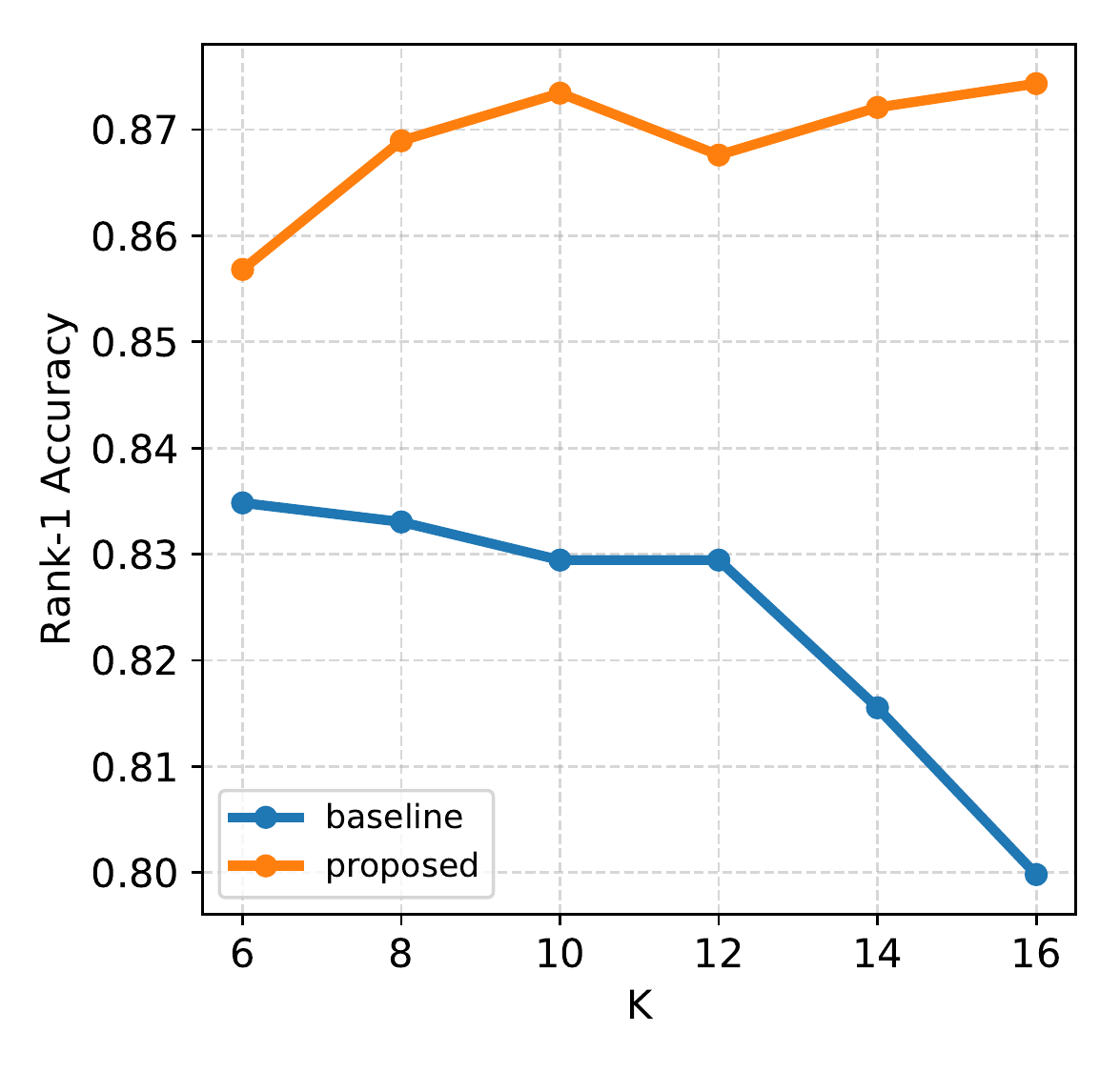}
	}
	\subfigure[MSMT7]{
		\label{fig:batch:b}
		\includegraphics[scale=0.345]{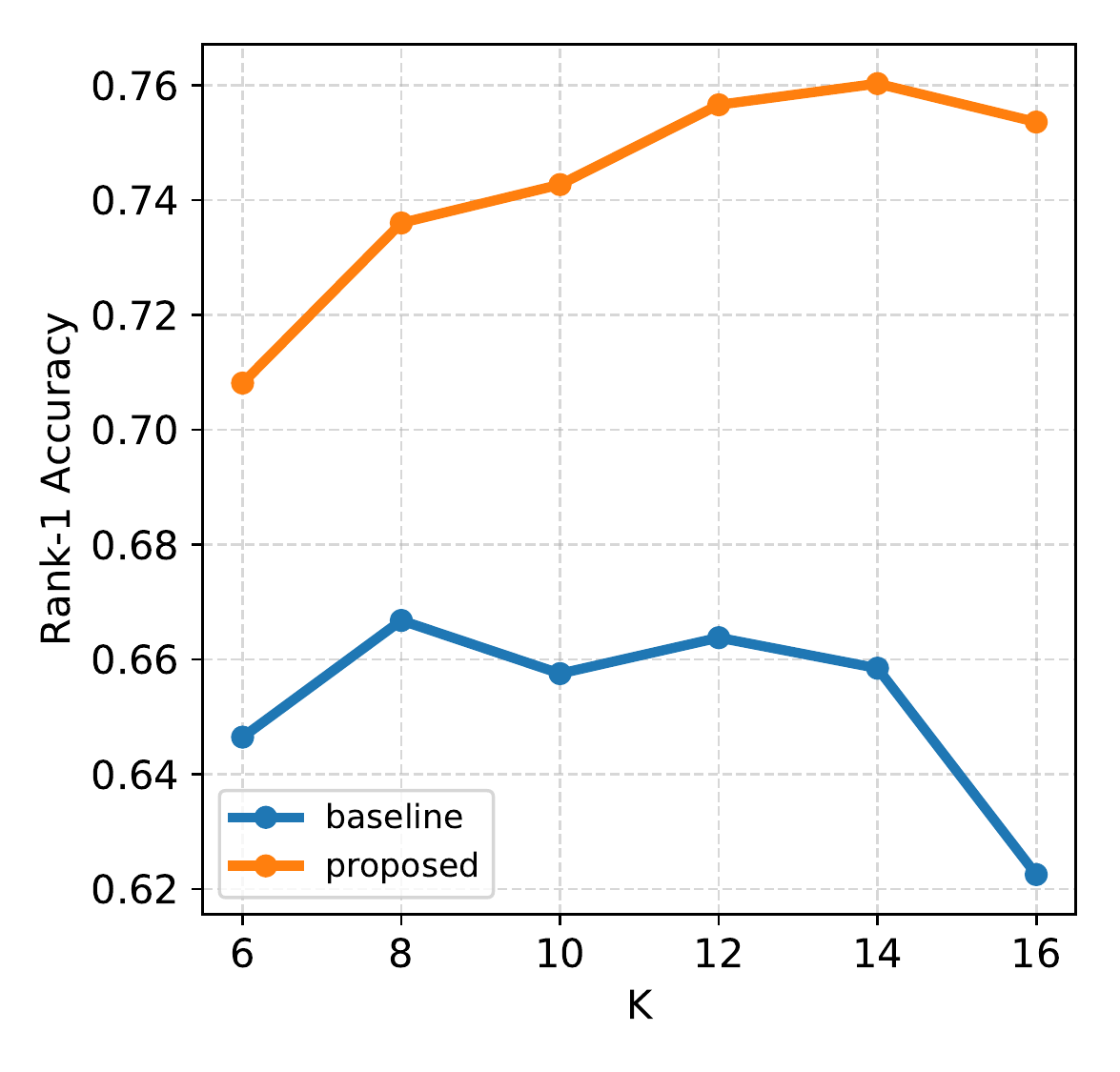}
	}
	\caption{The trend of performance while $K$(\#images per identity) varies. }
	\label{fig:batch}
\end{figure}

\noindent
{\bf Influence of The Number of Images per Identity $K$.}
We investigate the trend of the performance when varying $K$.
Given that Market-1501 and CUHK03 are relatively small which can not satisfy the need of larger $K$.
We only conduct experiments on MSMT17 and DukeMTMC-ReID.
Figure~\ref{fig:batch} shows that our approach can benefit from larger $K$, while the performance of vanilla baseline model even degrades dramatically when $K$ increases. This phenomenon again validates our hypothesis that group-wise training are more advantageous with larger mini-batch since for the training of a single sample, we can utilize the information of all the samples within the mini-batch.


\noindent
{\bf Comparison with Ncut Loss.}
As mentioned above, Ncut loss and our method realize the same idea from different aspects.
A comparison is thus performed between the two methods.
To make it fair, we implement Ncut loss based on the same backbone.
The results are summarized in Table~\ref{tab:nc}.
It is clear that our method outperforms Ncut loss on all benchmarks.
We conjecture that Ncut loss suffers from the detached optimization for feature and similarity.
While our method benefits a lot from optimizing feature and similarity jointly with a consistent objective. 

\begin{table}[h]
	\begin{center}
		\begin{tabular}{c|cc|cc}
			\Xhline{1.1px}
			\multirow{2}*{Dataset}& \multicolumn{2}{c|}{Ncut loss} & \multicolumn{2}{c}{ours} \\
			\cline{2-5}
			~  & mAP & R-1 & mAP & R-1 \\
			\hline
			Market-1501  & 79.5 & 92.0 & 82.7 & 93.4 \\
			DukeMTMC  & 66.7 & 84.0 & 73.2 & 86.9 \\ 
			CUHK03  & 40.3 & 45.5 & 62.4 & 68.2 \\
			MSMT17 & 37.4 & 66.3 & 47.6 & 73.6\\
			\Xhline{1.1px}
		\end{tabular}
	\end{center}
	\caption{Comparison between our method and Ncut.}
	\label{tab:nc}
\end{table}

\subsection{Comparison with State-of-the-art Methods}
The proposed method is compared with state-of-the-art methods in this section.  
To make it fair, we compare results with and without post-processing, respectively. 

\noindent
{\bf Results on Market-1501 dataset.}
As shown in Table~\ref{tab:market}, our method outperforms all competitors without tailor-made architectures or extra auxiliary information with only a single loss. 
We further perform a comparison on the dataset with 500k distractors.
The results are summarized in Table~\ref{tab:distractor}.
As reported in the table, our method is robust to distractors.
Specifically, when disturbed by 100k distractors, the mAP/rank-1 accuracy of our method only decreases by 4.9\%/2.5\%. 
Note that the rank-1 accuracy is still over 90\% in this case.
While for the other four competitors, the degradations are much larger than ours.
The performance gaps are even more significant when increasing the distractor size.

\begin{table}[h]
	\begin{center}
		\
		\begin{tabular}{l|c|ccc}
			\Xhline{1.1px}
			\multirow{2}*{Methods} & \multirow{2}*{Reference} & \multicolumn{3}{c}{Market-1501} \\
			~ & ~ & mAP & R-1 & R-5 \\
			\hline
			GLAD~\cite{wei2017glad} & ACMMM17 & 73.9 & 89.9 & - \\
			MLFN~\cite{chang2018multi} & CVPR18 & 74.3 & 90.0 & - \\
			HA-CNN~\cite{li2018harmonious} & CVPR18 & 75.7 & 91.2 & - \\
			DuATM~\cite{si2018dual} & CVPR18 & 76.6 & 91.4 & 97.1\\
			Part-aligned~\cite{suh2018part} & ECCV18  & 79.6  & 91.7 & 96.9\\
			PCB~\cite{sun2017beyond} & ECCV18 & 77.4 & 92.3 & 97.2\\
			Mancs~\cite{Wang_2018_ECCV} & ECCV18 & 82.3 & 93.1 & - \\
			Proposed & - & \textbf{82.7} & \textbf{93.4} & \textbf{97.4}\\
			\hline \hline
			GSRW~\cite{shen2018deep} & CVPR18 & 82.5 & 92.7 & 96.9 \\
			SGGNN~\cite{shen2018person} & ECCV18 & 82.8 & 92.3 & 96.1 \\
			Part-aligned(KR) & ECCV18 & 89.9 & 93.4 & 96.4 \\
			Proposed(R) & - & 87.5 & \textbf{94.1} & \textbf{97.5}\\
			Proposed(KR) & - & \textbf{90.6} & 93.5 & 96.6\\
			\Xhline{1.1px}
		\end{tabular}
	\end{center}
	\caption{Comparison with state-of-the-art methods on the Market-1501 dataset. (R) and (KR) means the model is refined by the proposed post-processing and k-reciprocal, respectively.}
	\label{tab:market}
\end{table}


\begin{table*}[h]
	\begin{center}
		\begin{tabular}{l|cc|cc|cc|cc}
			\Xhline{1.1px}
			\multirow{3}*{Methods} & \multicolumn{8}{c}{Distractor Size}  \\
			\cline{2-9}
			~  & \multicolumn{2}{c|}{0} & \multicolumn{2}{c|}{100k} & \multicolumn{2}{c|}{200k} & \multicolumn{2}{c}{500k} \\
			\cline{2-9}
			~ & mAP & Rank-1 & mAP & Rank-1 & mAP & Rank-1 & mAP & Rank-1 \\
\hline
			Zheng \etal~\cite{zheng2017discriminatively} & 59.9 & 79.5 & \dec{52.3}{7.6}& \dec{73.8}{5.7} & \dec{49.1}{10.8} & \dec{71.5}{8.0} & \dec{45.2}{14.7}  & \dec{68.3}{11.2} \\
			APR~\cite{lin2017improving} & 62.8 &  84.0 & \dec{56.5}{6.3} & \dec{79.9}{4.1} & \dec{53.6}{~9.2} & \dec{78.2}{5.8} & \dec{49.8}{13.0} & \dec{75.4}{~8.6} \\
			TriNet~\cite{hermans2017defense} & 69.1 & 84.9 & \dec{61.9}{7.2} & \dec{79.7}{5.2} & \dec{58.7}{10.4} & \dec{77.9}{7.0} & \dec{53.6}{15.5} & \dec{74.7}{10.2}\\
			Part-aligned~\cite{suh2018part} & 79.6 & 91.7 & \dec{74.2}{5.4} & \dec{88.3}{3.4} & \dec{71.5}{~8.1} & \dec{86.6}{5.1} & \dec{67.2}{12.4} & \dec{84.1}{~7.6}\\
			Proposed & 82.7 & 93.4 & \dec{77.8}{4.9} & \dec{90.9}{2.5} & \dec{75.5}{~7.2} & \dec{89.3}{4.1} & \dec{71.9}{10.8} & \dec{87.1}{~6.3} \\
			\Xhline{1.1px}
		\end{tabular}
	\end{center}
	\caption{Comparison with state-of-the-art methods on the Market-1501+500k dataset.}
	\label{tab:distractor}
\end{table*}

\noindent
{\bf Results on DukeMTMC-ReID dataset.}
The results on DukeMTMC-ReID dataset are presented in Table~\ref{tab:duke}. 
It can be seen that our method outperforms other state-of-the-arts significantly. 
Specifically, our approach gains 1.4\% and 2\% improvement over Mancs~\cite{Wang_2018_ECCV} in terms of mAP and rank-1 accuracy, respectively.

\begin{table}
	\begin{center}
		\begin{tabular}{l|c|ccc}
			\Xhline{1.1px}
			\multirow{2}*{Methods} & \multirow{2}*{Reference} & \multicolumn{3}{c}{DukeMTMC} \\
			~ & ~ & mAP & R-1 & R-5 \\
\hline
			PSE~\cite{sarfraz2017pose} & CVPR18 & 62.0 & 79.8 & 89.7 \\
			HA-CNN~\cite{li2018harmonious} & CVPR18 & 63.8 & 80.5 & - \\
			MLFN~\cite{chang2018multi} & CVPR18 & 62.8 & 81.0  & - \\
			DuATM~\cite{si2018dual} & CVPR18 & 64.6 & 81.8 & 90.2 \\ 			
			PCB+RPP~\cite{sun2017beyond} & ECCV18 & 69.2 & 83.3 & -\\
			Part-aligned~\cite{suh2018part} & ECCV18  & 69.3  & 84.4 & 92.2\\
			Mancs~\cite{Wang_2018_ECCV} & ECCV18 & 71.8 & 84.9 & - \\
			Proposed & - & \textbf{73.2} & \textbf{86.9} & \textbf{93.9}\\
\hline \midrule
			GSRW~\cite{shen2018deep} & CVPR18 & 66.4 & 80.7 & 88.5 \\
			SGGNN~\cite{shen2018person} & ECCV18 & 68.2 & 81.1 & 88.4 \\
			Part-aligned(KR) & ECCV18 & \textbf{83.9} & 88.3 & 93.1 \\
			Proposed(R) & - & 79.6 & \textbf{90.0} & \textbf{94.0}\\
			Proposed(KR) & - & 83.3 & 88.3 & 92.0\\
			\Xhline{1.1px}
		\end{tabular}
	\end{center}
	\caption{Comparison with state-of-the-art methods on the DukeMTMC-ReID dataset. (R) and (KR) means the model is refined by the proposed post-processing and k-reciprocal, respectively.}
	\label{tab:duke}
\end{table}

\noindent
{\bf Results on CUHK03 dataset.}
We only conduct experiments on the manually labeled subset of CUHK03. 
The results are reported in Table~\ref{tab:cuhk}.
It can be observed that our method achieves best performance among compared methods.

\begin{table}
	\begin{center}
		\begin{tabular}{l|c|ccc}
			\Xhline{1.1px}
			\multirow{2}*{Methods} & \multirow{2}*{Reference} & \multicolumn{3}{c}{CUHK03} \\
			~ & ~ & mAP & R-1 & R-5 \\
\hline
			SVDNet~\cite{Sun2017SVDNet} & ICCV17& 37.8 & 40.9 & - \\
			DPFL~\cite{chen2017person}& ICCV17 & 40.5 & 43.0 & - \\
			HA-CNN~\cite{li2018harmonious} & CVPR18 & 41.0 & 44.4 & - \\
			MLFN~\cite{chang2018multi} & CVPR18 & 49.2 & 54.7 & - \\
			DaRe~\cite{wang2018resource} & CVPR18 & 61.6 & 66.1 & - \\
			Proposed & - & \textbf{62.4} & \textbf{68.2} & \textbf{84.4}\\
			\Xhline{1.1px}
		\end{tabular}
	\end{center}
	\caption{Comparison with state-of-the-art methods on the CUHK03 dataset. We adhere to newly proposed evaluation protocol~\cite{zhong2017re} and report results on manually labeled version of CUHK03. }
	\label{tab:cuhk}
\end{table}

\noindent
{\bf Results on MSMT17 Dataset.}  
Since MSMT17 is released very recently, there is no other published work evaluated on it to our best knowledge. 
So we only compare our method with baselines reported by authors~\cite{wei2018person}.
As shown in Table~\ref{tab:msmt}, our method outperforms these baselines dramatically.
Specifically, it exceeds GLAD by 13.6\% and 12.2\% in terms of mAP and rank-1 accuracy, respectively. 
This verifies the scalability and the robustness of our method when applied in large scale scenarios.   
To clarify the superiority of our method, we remind readers that GLAD~\cite{wei2017glad} performs pretty well on Market-1501 as recorded in Table~\ref{tab:market}.

\begin{table}
	\begin{center}
		\begin{tabular}{l|c|ccc}
			\Xhline{1.1px}
			\multirow{2}*{Methods} & \multirow{2}*{Reference} & \multicolumn{3}{c}{MSMT17} \\
			~ & ~ & mAP & R-1 & R-5 \\
\hline
			GoogleNet~\cite{wei2018person} & CVPR18 & 23.0 & 47.6 & 65.0 \\
			PDC~\cite{wei2018person} & CVPR18 & 29.7 & 58.0 & 73.6\\
			GLAD~\cite{wei2018person} & CVPR18 & 34.0 & 61.4 & 76.8\\
			Proposed & - & \textbf{47.6} & \textbf{73.6} & \textbf{85.6}\\
			\Xhline{1.1px}
		\end{tabular}
	\end{center}
	\caption{Comparison with state-of-the-art methods on the MSMT17 dataset.}
	\label{tab:msmt}
\end{table}

\section{Conclusion}
\label{sec:conclusion} 
Inspired by classical spectral clustering, we have proposed a novel spectral feature transformation module to facilitate the learning of discriminative features.
In contrast to most of other methods that process samples individually, our method defines a group-wise loss function by the spirit of spectral clustering, and then optimize the deep neural networks by the guidance of this novel loss.
The module only involves several basic matrix operations, but the improvement it brings is significant.
Furthermore, we extend it to post-processing which effectively improves the top ranking of the results.
Ablation studies on four benchmarks prove the effectiveness and scalability of our method.
It reinforces the strong baseline significantly and outperforms other state-of-the-art without bells and whistles.

{\small
\bibliographystyle{ieee}
\bibliography{egbib}
}

\end{document}